\pdfoutput=1
\documentclass[11pt]{article}

\usepackage{acl}

\usepackage{graphicx}
\usepackage{amsmath}
\usepackage{mathtools}
\usepackage{amssymb}
\usepackage{times}
\usepackage{latexsym}
\usepackage{booktabs}
\usepackage{multirow}
\usepackage{adjustbox}
\usepackage{siunitx}
\usepackage{tabularx}

\usepackage{array}

\usepackage{algorithm}
\usepackage{algpseudocode}

\usepackage[T1]{fontenc}
\usepackage[utf8]{inputenc}

\usepackage{microtype}

\usepackage{array}
\newcolumntype{P}[1]{>{\centering\arraybackslash}p{#1}}
\newcolumntype{L}[1]{>{\raggedright\arraybackslash}p{#1}}

\newcolumntype{F}[1]{%
    >{\raggedright\arraybackslash\hspace{0pt}}p{#1}}%
\newcolumntype{T}[1]{%
    >{\centering\arraybackslash\hspace{0pt}}p{#1}}%

\newcolumntype{C}[1]{>{\centering\arraybackslash}p{#1}}

\title{Exploring the Effectiveness of Instruction Tuning in Biomedical Language Processing}

\author{
  Omid Rohanian$^{1,2\footnotemark[1]}$,
  Mohammadmahdi Nouriborji$^{2,4\footnotemark[1]}$, 
  \textbf{David A. Clifton}$^{1,3}$\\
  $^1$Department of Engineering Science, University of Oxford, Oxford, UK \\
  $^2$NLPie Research, Oxford, UK \\
  $^3$Oxford-Suzhou Centre for Advanced Research, Suzhou, China \\
  $^4$Sharif University of Technology, Tehran, Iran \\
  \texttt{\{m.nouriborji,omid\}@nlpie.com}\\
  \texttt{\{omid.rohanian,david.clifton\}@eng.ox.ac.uk}\\
}

\begin{document}
\maketitle

\renewcommand*{\thefootnote}{\fnsymbol{footnote}}
\footnotetext[1]{Both authors contributed equally to this work.}

\renewcommand*{\thefootnote}{\arabic{footnote}}
\setcounter{footnote}{0}

\begin{abstract}
Large Language Models (LLMs), particularly those similar to ChatGPT, have significantly influenced the field of Natural Language Processing (NLP). While these models excel in general language tasks, their performance in domain-specific downstream tasks such as biomedical and clinical Named Entity Recognition (NER), Relation Extraction (RE), and Medical Natural Language Inference (NLI) is still evolving. In this context, our study investigates the potential of instruction tuning for biomedical language processing, applying this technique to two general LLMs of substantial scale. We present a comprehensive, instruction-based model trained on a dataset that consists of approximately $200,000$ instruction-focused samples. This dataset represents a carefully curated compilation of existing data, meticulously adapted and reformatted to align with the specific requirements of our instruction-based tasks. This initiative represents an important step in utilising such models to achieve results on par with specialised encoder-only models like BioBERT and BioClinicalBERT for various classical biomedical NLP tasks. Our work includes an analysis of the dataset's composition and its impact on model performance, providing insights into the intricacies of instruction tuning. By sharing our codes, models, and the distinctively assembled instruction-based dataset, we seek to encourage ongoing research and development in this area.
\end{abstract}

\section{Introduction}

Transformers have become the cornerstone of modern NLP, providing the backbone for a wide array of applications including machine translation, question-answering, and text summarisation \citep{vaswani2017attention}. Their self-attention mechanisms and parallelised architecture have proven to be highly effective in capturing the nuances of human language \citep{devlin-etal-2019-bert}.

Autoregressive language models, exemplified by the Generative Pre-trained Transformer series like GPT \citep{radford2018improving} and GPT-3 \citep{brown2020language}, have revolutionised the way NLP is approached. These models, operating as decoder-only transformers, excel at generating text in a sequential, token-by-token manner, leveraging their attention mechanisms to focus on relevant segments of input text. Models based on this architecture, such as GPT-4 have demonstrated a remarkable ability to perform a variety of language tasks without the need for task-specific fine-tuning, showcasing strong zero-shot and few-shot learning capabilities. This feature allows these models to effectively respond to text-based prompts, including those with a limited number of examples or instructions, thereby enabling a more interactive and dynamic text generation process.

Medical language models, particularly encoder-only models like BioBERT and ClinicalBERT, have been instrumental in advancing tasks such as medical diagnosis, biomedical literature mining, and clinical information extraction \citep{clusmann2023future,kormilitzin2021med7}. Excelling in areas like classification and Named Entity Recognition (NER), these models have significantly contributed to biomedical NLP. However, they often lack inherent capabilities in interpreting and executing natural language instructions or generating reports from medical Electronic Health Records (EHRs). This limitation has spurred research into developing generative Large Language Models (LLMs) capable of handling more dynamic tasks, aiming to parallel the performance of specialised encoder-only models in the biomedical domain. Yet, as indicated by studies such as Lehman et al. \citep{pmlr-v209-eric23a}, encoder-only models continue to lead in clinical NLP, underscoring the challenges in tailoring general-domain LLMs for specialised medical applications. Our research aims to contribute to this area by introducing a dataset that integrates various clinical and biomedical datasets. Utilising this resource, we apply instruction tuning to two publicly available general LLMs, with the objective of exploring its potential in enhancing the performance of these LLMs for downstream medical tasks. This approach represents an initial step towards understanding the effectiveness of instruction tuning in this domain, with the dataset serving as an additional tool to facilitate this exploration for future work.

% Our contributions
The primary contributions of our work are as follows. First, we introduce Llama2-MedTuned, developed in two variants: one fine-tuned on the Llama2 7B model\footnote{Llama2-MedTuned-7b: \url{https://huggingface.co/nlpie/Llama2-MedTuned-7b}.} and the other on the Llama2 13B model\footnote{Llama2-MedTuned-13b: \url{https://huggingface.co/nlpie/Llama2-MedTuned-13b}.}. These are specialised models designed explicitly for instruction-based tasks in the medical domains. Second, we present a dataset that amalgamates various publicly available datasets into a novel configuration, creating a rich and diverse training environment specifically compiled for the Llama2-MedTuned models. Our comparative experimental results highlight the effectiveness of our approach in comparison to current state-of-the-art models in a number of classical tasks in biomedical and clinical NLP.

\section{Related Works}
\label{related}

\begin{figure*}[t!]
\centering
\includegraphics[scale=0.5]{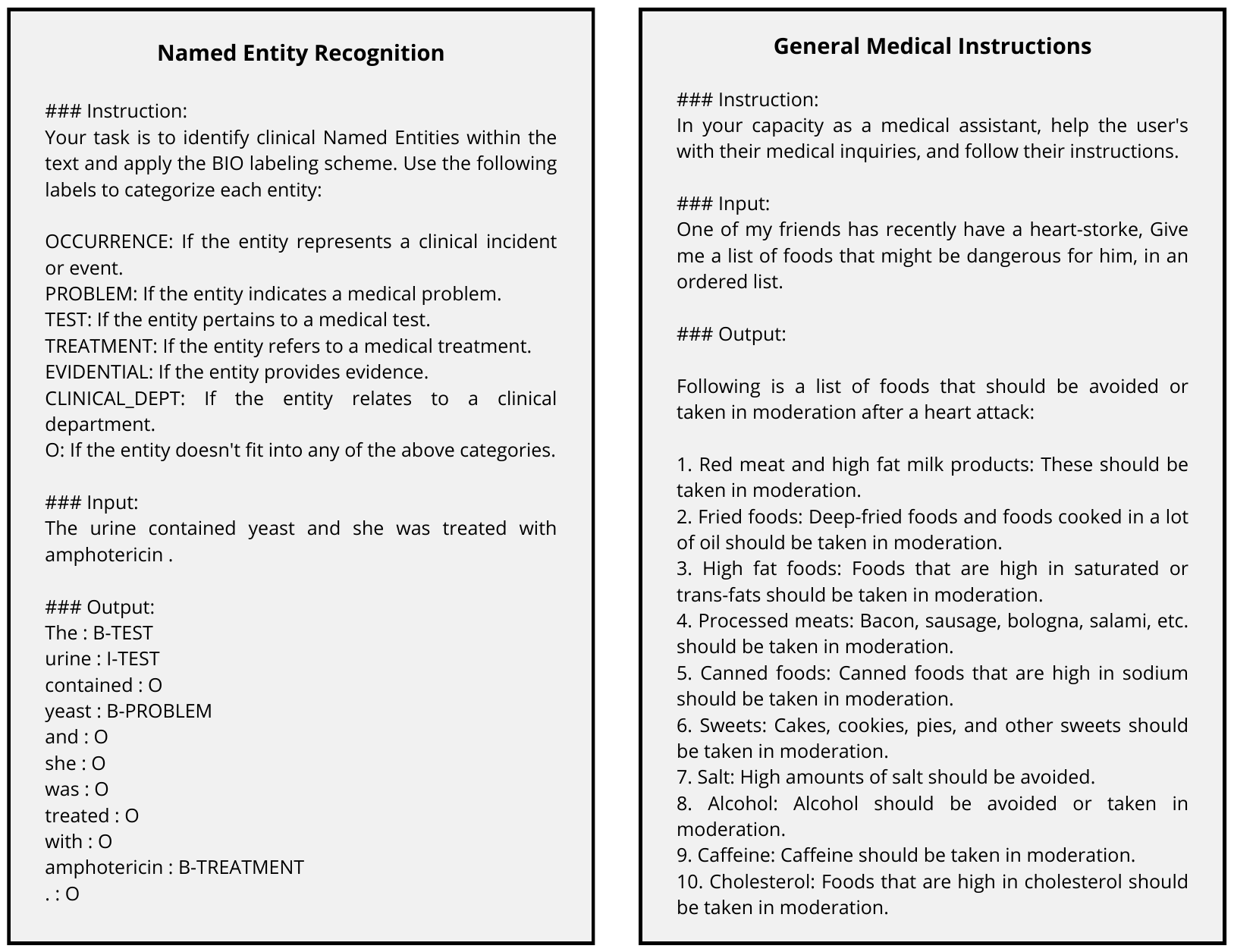}
\caption{Example outputs from Llama2-MedTuned-7B for biomedical tasks (left) and general medical instructions (right). The model demonstrates the application of instruction-based learning in NER by correctly labeling biomedical entities (left) and providing a relevant list in response to a medical inquiry (right).}
\label{fig:generation-samples}
\end{figure*}

\subsection{Autoregressive Language Models}
Autoregressive Language Models (ALMs), exemplified by GPT and its different variants, constitute a class of transformers pre-trained on a language modelling objective, namely, predicting the subsequent token given a particular context \citep{radford2018improving,radford2019language}. Noteworthy instances of ALMs include GPT-3.5 and GPT-4 by OpenAI, trained on extensive datasets harvested from the web for the language modelling objective \citep{brown2020language}. Google's Bard/Gemini and Anthropic's Claude are also notable contributions to this field, demonstrating the growing exploration and advancements of autoregressive language models for diverse applications.

\subsection{Instruction-Based Language Models}
Instruction-based language models, a novel category within autoregressive models, have been shown to improve significantly when fine-tuned with instructions. Traditional autoregressive models, while adept at sequential text generation, often struggle with comprehending and executing complex instructions. Fine-tuning such models on natural language instructions and human-generated responses can markedly enhance their ability to follow instructions accurately \citep{wei2021finetuned}. This advancement is exemplified in models like Instruct-GPT \citep{ouyang2022training}, Falcon \citep{refinedweb}, and Llama \citep{touvron2023llama}, which are fine-tuned to respond more effectively to instruction-based prompts, thus enabling more dynamic and interactive text generation capabilities.

\subsection{Clinical LLMs}
With the advent of instruction-based LLMs, their adaptation to the clinical domain has been explored, using instruction-based datasets specific to this area. ChatDoctor \citep{li2023chatdoctor}, a fine-tuned clinical chatbot, has been trained on real conversations between doctors and patients, showcasing its efficacy in clinical settings. Similarly, Med-Alpaca \citep{han2023medalpaca} and Clinical Camel \citep{toma2023clinical} follow this trend by adapting open LLMs to the clinical domain. PMC-Llama \citep{wu2023pmcllama} is another significant model, initially pre-trained on a biomedical/clinical corpus, and subsequently trained on an instruction dataset primarily containing medical question answering and reasoning tasks.

\section{Method}
\label{method}

\begin{figure*}[t!]
\centering
\includegraphics[scale=0.38]{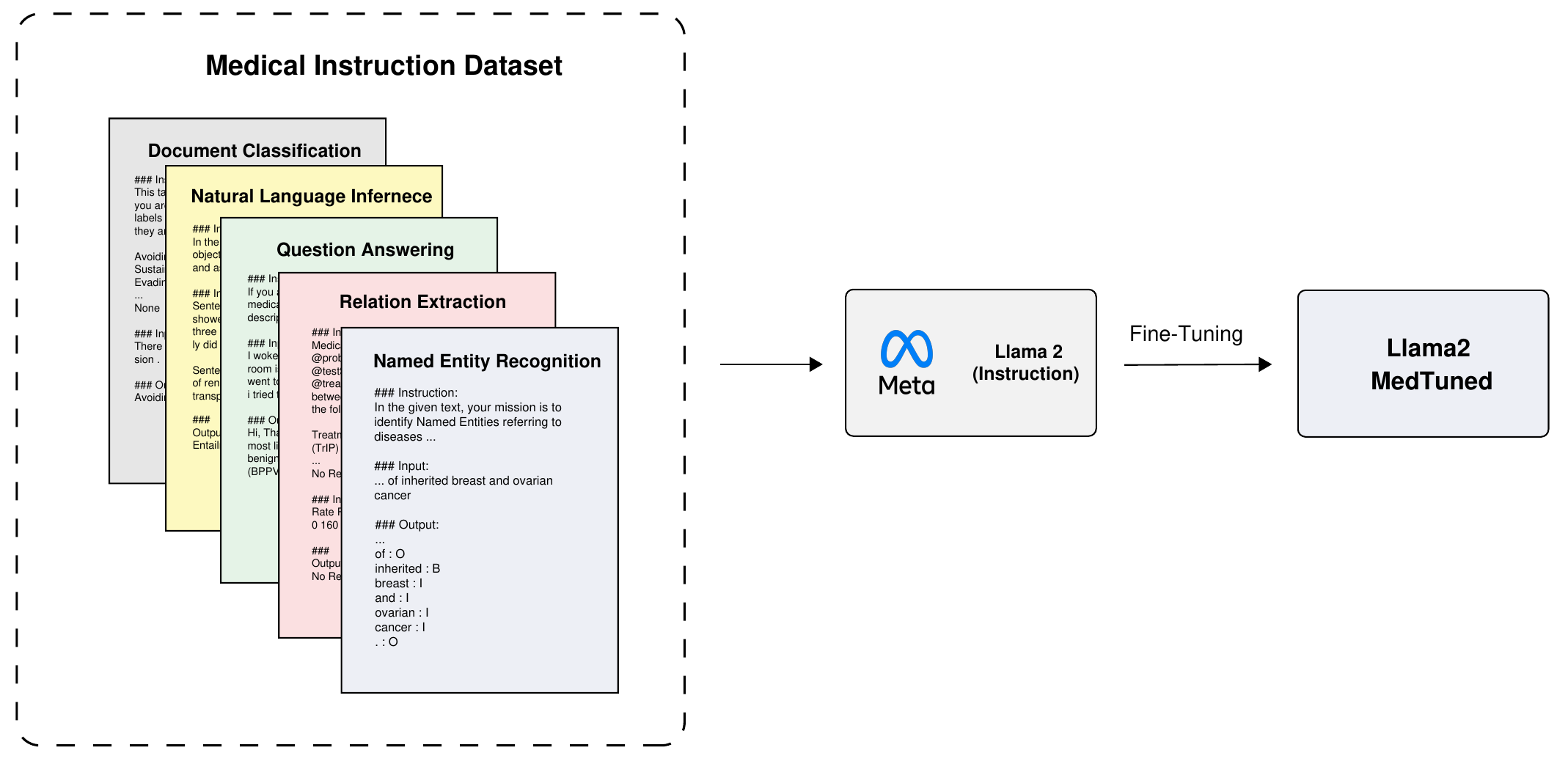}
\caption{Schematic representation of the process for fine-tuning Llama2 models with the proposed medical instruction dataset.}
\label{fig:architecture}
\end{figure*}

\begin{figure*}[h!]
\centering
\includegraphics[scale=0.6]{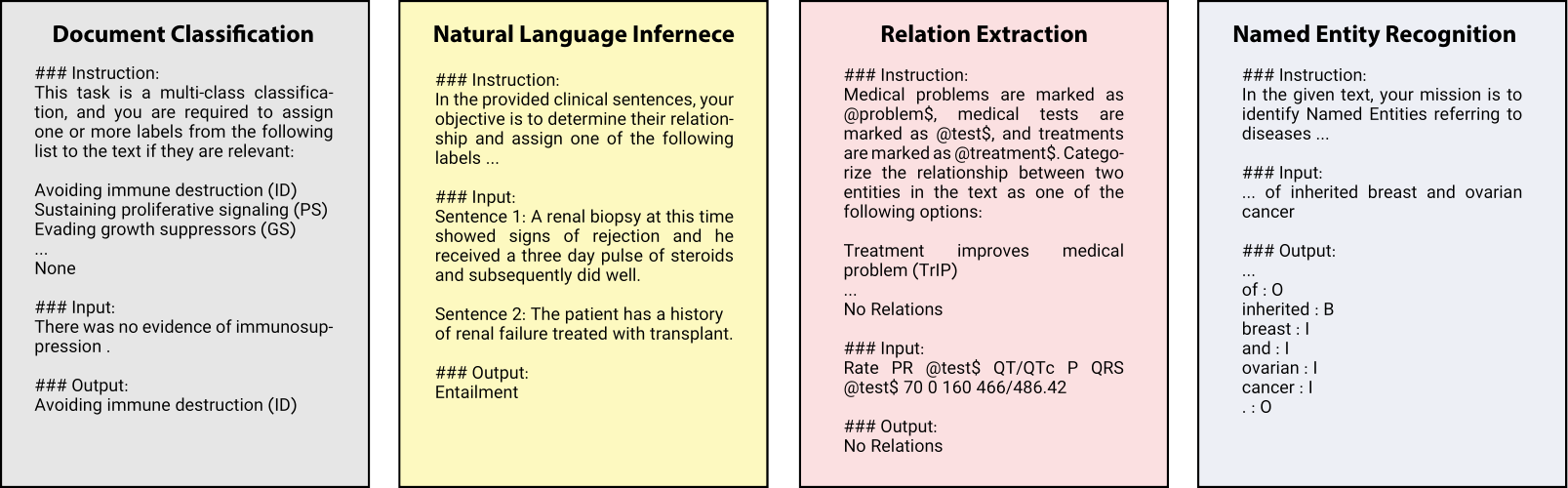}
\caption{Overview of some of the prompt templates used in our instruction dataset.}
\label{fig:prompts-sample}
\end{figure*}

In this work, we train an instruction-based language model for the medical domain which is able to target tasks such as Named Entity Recognition, Relation Extraction, Document Classification, Question Answering, and Natural Language Inference. In order to train this model, we compiled a new medical instruction-based dataset called \textbf{Llama2-MedTuned-Instructions}\footnote{\url{https://huggingface.co/datasets/nlpie/Llama2-MedTuned-Instructions}}.

\subsection{Prompting Template}
To transform the original datasets into instruction-based formats, we adopted the prompting strategy used in the Alpaca dataset. Our prompts are composed of three parts: Instruction, Input, and Output. In the Instruction section, we developed $5$ to $10$ different instructions for each dataset, detailing the target tasks and the labelling scheme for the model. One instruction is randomly chosen for each sample during the conversion to the instruction-based dataset. The Input is the dataset's original input, while the Output is the expected output that the model should predict, consistent with the format described in the instructions. Figure \ref{fig:prompts-sample} presents some samples from our instruction dataset.

\subsection{Tasks and Datasets}
As mentioned earlier, various tasks are used in this work to diversify the training corpus used for training our language model. Training subsets from several well-known datasets were selected for each task to assemble the dataset employed in our study.

\subsubsection{Named Entity Recognition}
For the task of Named Entity Recognition, we used the NCBI-disease, BC5CDR-disease \citep{dougan2014ncbi}, BC5CDR-chem \citep{li2016biocreative}, BC2GM \citep{smith2008overview}, JNLPBA \citep{smith2008overview}, and i2b2-2012 dataset \citep{sun2013evaluating}. For the first five datasets, we use the BIO labeling scheme with no additional label names. However, for the i2b2-2012 dataset, $6$ different categories are used along with BIO labeling.

\subsubsection{Relation Extraction} 
We used the i2b2-2010 \citep{uzuner20112010} and GAD \citep{bravo2015extraction} datasets for relation extraction. For both datasets we follow the same pre-processing method used in \citet{10.1093/bioinformatics/btad103} and \citet{rohanian2023lightweight}, which uses specific tags (e.g. \@test\$, \@problem\$, etc.) for tagging medical concepts in the text, in order to frame the relation extraction as a sentence classification task.

\subsubsection{Natural Language Inference}
For Natural Language Inference, we used the MedNLI dataset \citep{romanov2018lessons}, which is composed of pairs of medical sentences labeled with Entailment, Contradiction, or Neutral to indicate the type of relationship between them.

\subsubsection{Document Classification}
We used the hallmarks of cancer (HoC) dataset \citep{baker2015automatic} for the task of Document Classification which is a well-known multi-class classification dataset in the medical domain.

\subsubsection{Question Answering}
For question answering, we used two prominent datasets, ChatDoctor \citep{li2023chatdoctor}, and Pmc-Llama-Instructions \citep{wu2023pmcllama}. ChatDoctor consists of $100$k samples taken from the ChatDoctor website that are real conversations between patients and doctors, In our dataset we randomly sampled 50K samples from this dataset. PMC-Llama-Instructions is a large dataset consisting of multiple QA datasets such as MedQA \citep{jin2021disease}, PubMedQA \citep{jin2019pubmedqa}, etc. For our work, we randomly sampled 50K samples from this dataset.

\subsubsection{Llama2-MedTuned Instructions}
Finally, we concatenate all of the datasets mentioned earlier in this section and shuffle them to obtain our final medical instruction-based dataset which consists of approximately $200$K samples.

\subsection{Training Configuration}
In order to train our models, we used $10$ V100 GPUs with a batch size of $4$ per GPU. We used the deepspeed zero 3 config without CPU offloading, with a learning rate of $1e-5$ and $500$ warmup steps along with a linear learning rate scheduler. The models were trained for three epochs.

\begin{table*}[ht!]
    \centering
    \small
    \caption{Ablation study results using the instruction-based dataset. The $\dagger$ symbol denotes the model trained with the expanded instruction dataset.}
    \label{ablation-table}
    \setlength{\tabcolsep}{4pt}  % Adjusts the space between columns
    \renewcommand{\arraystretch}{1.2}  % Adjusts the space between rows
    \begin{tabular}{L{2cm}L{3cm}C{2cm}C{2cm}}
        \toprule[1pt]
        Type & Task & Llama2-MedTuned & Llama2-MedTuned$^{\dagger}$ \\\midrule[0.5pt]
        NER & NCBI-Disease & 85.69 & 83.59\\
        NER & BC5CDR-Disease & 85.46 & 84.30\\
        NER & BC5CDR-Chem & 94.51 & 93.77\\
        NER & BC2GM & 79.12 & 78.51\\
        NER & JNLPBA & 81.31 & 78.91\\
        \bottomrule
    \end{tabular}
\end{table*}

\section{Results}
\label{results}

Assessing the instruct-tuned models, Llama2-MedTuned 7B and 13B, against their foundational counterparts, Llama2 7B and 13B, presents complexities. As depicted in Figures \ref{fig:ner-outputs} and \ref{fig:re-outputs} in the appendix, prompting the base Llama2 models for NER often yields outputs that are erratic and difficult to quantify. Our study is limited to zero-shot learning scenarios\footnote{One-shot and zero-shot cases resulted in virtually identical results, therefore we are only reporting the results of the zero-shot run.}, where tasks and expected outcomes are defined in basic terms. Although venturing into few-shot learning or advanced prompting might alter the results, our limited experimentation did not indicate a significant shift in this trend. Therefore, barring MedNLI, where Llama2 generates consistent outputs, our models are compared with DistilBERT and BioBERT, the conventional baselines for NER, RE, and NLI tasks.

Thanks to instruction-tuning, we were able to systematically interpret our models' outputs into a structured format, suitable for evaluation using conventional metrics like F1 or Accuracy. The results for the biomedical NER are available in Table \ref{t:bio-results}, Where the $13$B model is generally better than our $7$B model. Additionally, the results of the clinical tasks are available in Table \ref{t:med-results}. 

Generally, interpreting the outputs of Llama2 on most structured tasks proved to be challenging as the outputs tended to deviate from the expected format. We have provided examples of output generations from both our model and Llama2 in Figures \ref{fig:ner-outputs} and \ref{fig:re-outputs}. \citet{jahan2023comprehensive} reports results for the NER datasets on a number of closed and open LLMs including LLama2. Please refer to table \ref{tab:base-results} for a baseline reference to the reported results on the NER tasks in the literature. Llama2, on the other hand, did yield consistent outputs on the MedNLI task. Upon evaluation, the Llama2 model scored an accuracy of $37.20$ on the MedNLI evaluation subset, significantly lower than the $89.46$ score achieved by Llama2-MedTuned-13b.

\begin{table*}[ht!]
    \centering
    \small
    \caption{Test results on the Biomedical downstream tasks}
    \label{t:bio-results}
    \setlength{\tabcolsep}{4pt}  % Adjusts the space between columns
    \renewcommand{\arraystretch}{1.2}  % Adjusts the space between rows
    \begin{tabular}{L{2cm}L{3cm}C{2cm}C{2cm}C{2cm}C{2cm}}
        \toprule[1pt]
        Type & Task & DistilBERT & BioBERT-v1.1 & Llama2-MedTuned-7b & Llama2-MedTuned-13b \\\midrule[0.5pt]
        NER & NCBI-Disease & 86.38 & 88.62 & 87.18 & 85.69\\
        NER & BC5CDR-Disease & 82.01 & 86.67 & 83.92 & 85.46\\
        NER & BC5CDR-Chem & 92.50 & 94.73 & 93.88 & 94.51\\
        NER & BC2GM & 84.61 & 87.62 & 76.46 & 79.12\\
        NER & JNLPBA & 79.14 & 80.33 & 82.30 & 81.31\\
        \bottomrule
    \end{tabular}
\end{table*}

\begin{table*}[ht!]
    \centering
    \small
    \caption{Test results on the clinical downstream tasks}
    \label{t:med-results}
    \setlength{\tabcolsep}{4pt}  % Adjusts the space between columns
    \renewcommand{\arraystretch}{1.2}  % Adjusts the space between rows
    \begin{tabular}{L{2cm}L{2cm}C{2cm}C{3cm}C{2cm}C{2cm}}
        \toprule[1pt]
        Type & Task & DistilBERT & BioClinicalBERT & Llama2-MedTuned-7b & Llama2-MedTuned-13b \\\midrule[0.5pt]
        NER & i2b2-2012 & 79.15 & 82.98 & 80.67 & 80.64 \\
        RE  & i2b2-2010 & 92.75 & 93.58 & 89.35 & 93.14 \\
        NLI & MedNLI & 73.41 & 82.41 & 79.21 & 89.46 \\
        \bottomrule
    \end{tabular}
\end{table*}

\section{Ablation Studies}
To maintain the general capabilities of our model on tasks such as Question Answering and general instructions we use additional instruction-based data along with our NER, RE, and CLS instructions. We tested two strategies to create our final dataset. First, we randomly sampled $50$K samples from the PMC-Llama instructions, and $50$K from the ChatDoctor. For the second approach, we employed a more balanced sampling by taking $50$K samples from PubMedQA, $50$K from MedQA, $100$\% of UMLS relations, and UMLS, which resulted in $200$K samples from the PMC-Llama instructions, along with $50$K samples from  ChatDoctor. The ablation study results, presented in Table \ref{ablation-table}, reveal that the model trained on the larger PMC-Llama dataset exhibited inferior performance in biomedical downstream tasks compared to the model trained on the smaller dataset.

\footnotetext[1]{The results are taken from \citep{jahan2023comprehensive}}

\begin{table}[htbp]
  \centering
  \small
  \caption{Baseline results of different language models on the biomedical NER tasks \footnotemark[1]}
  \begin{tabular}{lcccc}
    \toprule
    \multirow{2}{*}{Dataset} & \multirow{2}{*}{GPT-3.5} & \multirow{2}{*}{Llama-2} & \multirow{2}{*}{Claude-2} \\\\
    \midrule
    NCBI-disease & 33.39 & 4.58 & 45.75 \\
    BC2GM & 31.99 & 5.95 & 40.45 \\
    BC5CDR-chem & 41.25 & 12.21 & 58.05 \\
    BC5CDR-disease & 32.26 & 5.68 & 50.13 \\
    JNLPBA & 31.89 & 4.30 & 34.62 \\
    \bottomrule
  \end{tabular}
  \label{tab:base-results}
\end{table}

\section{Conclusions \& Future Works}
In our study, we focused on instruction tuning of the Llama 2 model using a bespoke biomedical dataset, specifically curated for specialised biomedical NLP tasks like Named Entity Recognition (NER), Relation Extraction (RE), and medical Natural Language Inference (NLI). This process led to the creation of Llama2-MedTuned-7B and Llama2-MedTuned-13B, which represent adaptations of the original Llama 2 models. These tuned versions showed significant improvements in handling the complexities of medical NER, RE, and NLI indicating the efficacy of instruction tuning in aligning general-purpose language models with specialised task requirements. Despite these advancements, the scope for further enhancement remains, particularly in terms of processing detailed clinical instructions and further evaluating performance on Question Answering (QA) tasks. Future initiatives will involve enriching our dataset to cover an even broader range of biomedical and clinical tasks. Moreover, in line with the rapid advancements in the NLP field, we intend to explore and potentially integrate cutting-edge language models like Mistral, continually refining our approach to meet the dynamic needs of biomedical NLP applications.

\section*{Funding}
This work was supported in part by the National Institute for Health Research (NIHR) Oxford Biomedical Research Centre (BRC), and in part by an InnoHK Project at the Hong Kong Centre for Cerebro-cardiovascular Health Engineering (COCHE). OR acknowledges the support of the Medical Research Council (grant number MR/W01761X/). DAC was supported by an NIHR Research Professorship, an RAEng Research Chair, COCHE, and the Pandemic Sciences Institute at the University of Oxford. The views expressed are those of the authors and not necessarily those of the NHS, NIHR, MRC, COCHE, or the University of Oxford.

\section*{Limitations}
Our exploration into the application of large autoregressive language models like Llama2 for NLP tasks such as NER and RE unveiled significant challenges. The base Llama2 models, without fine-tuning, struggled to generate coherent and appropriately formatted outputs for these tasks. This underscores the difficulty in applying general-purpose LLMs to domain-specific tasks that demand organised responses. However, our instruct-tuned models, Llama2-MedTuned 7B and 13B, showed improved performance, successfully generating outputs in the necessary structured format. Despite this advancement, they did not outperform specialised models like BioBERT, highlighting a need for further development to meet the precision requirements of specific biomedical NLP tasks.

\bibliography{anthology,custom}
\bibliographystyle{acl_natbib}

\appendix

\section{Appendix}
\label{sec:appendix}

\begin{figure*}[h!]
\centering
\includegraphics[scale=0.8]{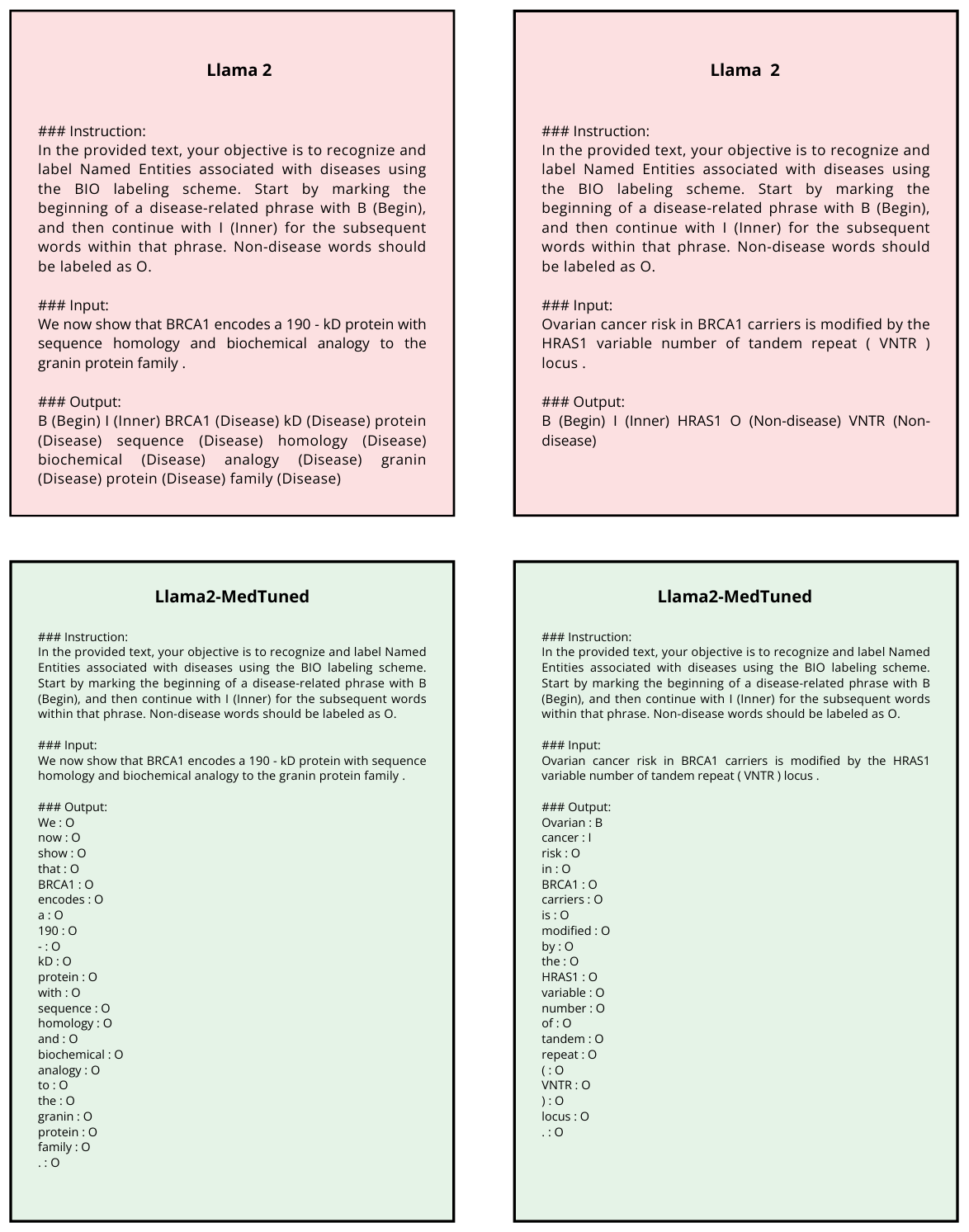}
\caption{Sample outputs of the Llama2 model and Llama2-MedTuned on Named Entity Recognition}
\label{fig:ner-outputs}
\end{figure*}

\begin{figure*}[h!]
\centering
\includegraphics[scale=0.8]{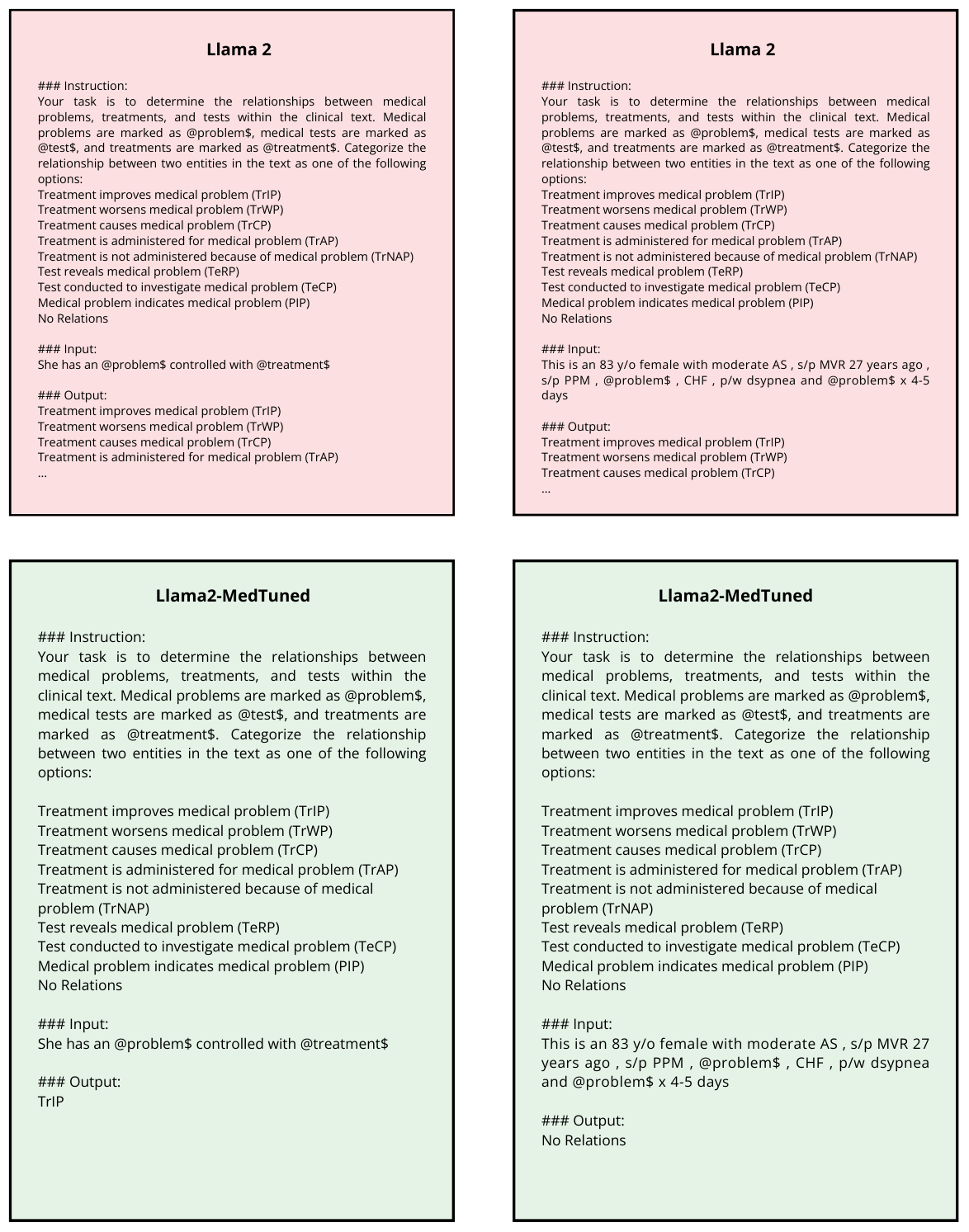}
\caption{Sample outputs of the Llama2 model and Llama2-MedTuned on Relation Extraction}
\label{fig:re-outputs}
\end{figure*}

\begin{figure*}[h!]
\centering
\includegraphics[scale=0.65]{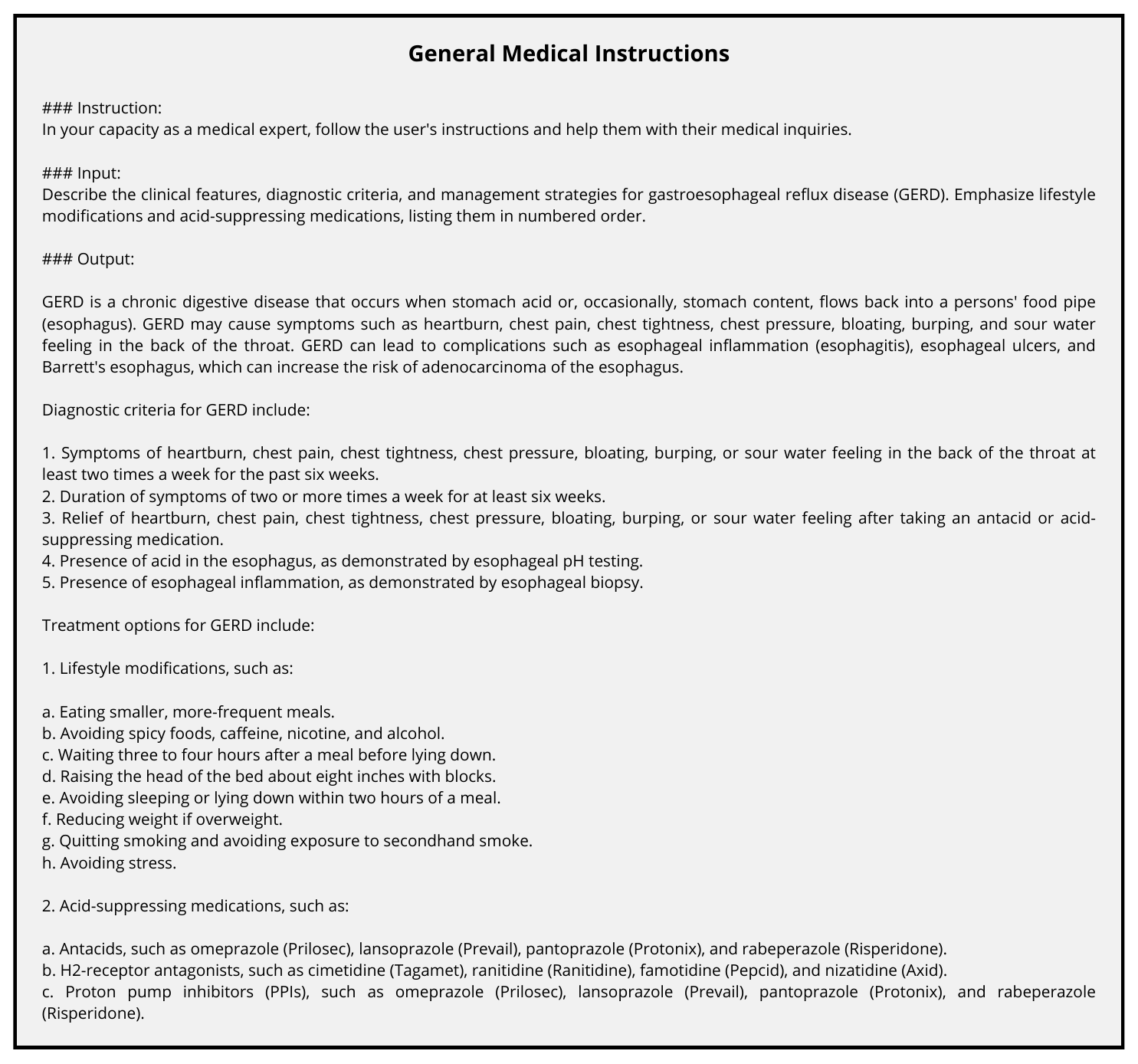}
\caption{Llama2-MedTuned output sample on general medical instructions}
\label{fig:prompts1}
\end{figure*}

\begin{figure*}[h!]
\centering
\includegraphics[scale=0.6]{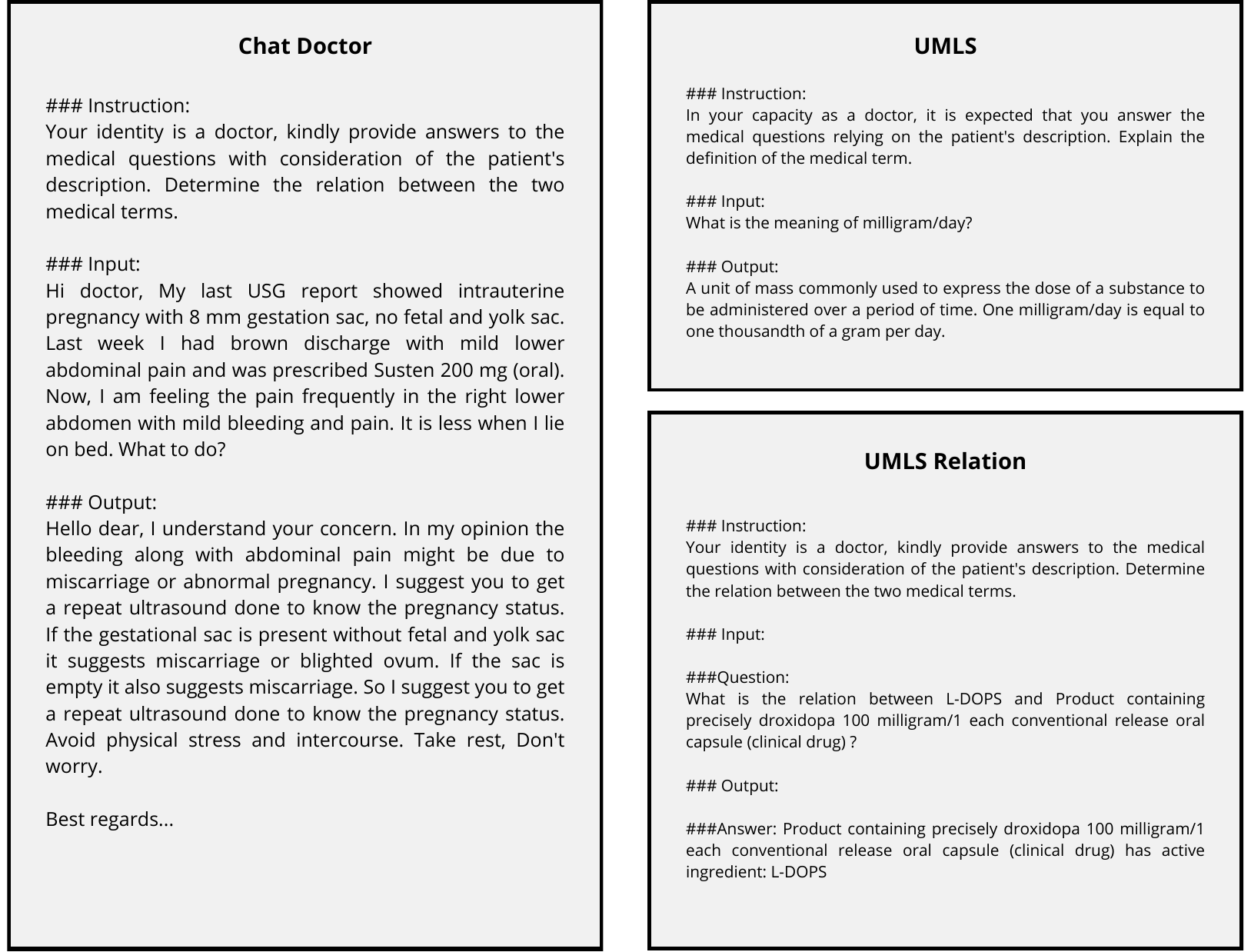}
\caption{Llama2-MedTuned outputs on a few medical generation tasks}
\label{fig:prompts2}
\end{figure*}

\end{document}